\documentclass{WileyMSP-template}
\usepackage{enumerate}
\usepackage{subfigure}
\usepackage{booktabs}
\usepackage{longtable}
\usepackage{amsmath,amssymb,amsfonts}%
\usepackage{xcolor}%

\begin{document}
\justifying

\pagestyle{fancy}

\title{A quantum moving target segmentation algorithm for grayscale video}

\maketitle


\author{Wenjie Liu*}
\author{Lu Wang}
\author{Qingshan Wu}



\begin{affiliations}
W. Liu, Q. Wu\\
School of Software, Nanjing University of Information Science  and Technology,
Nanjing, 210044, Jiangsu, China\\
Email Address:wenjiel@163.com

L. Wang\\
School of Information Science and Engineering, Southeast University, Nanjing, 211189, Jiangsu, China\\
Email Address:Lu\_Wang\_MT@163.com

\end{affiliations}


\keywords{Quantum video processing, QVNEQR, Moving target segmentation, Three-frame difference, IBM Q}

\begin{abstract}

The  moving target segmentation (MTS) aims to segment out moving targets in the video, however, the classical algorithm faces the huge challenge of real-time processing in the current video era. Some scholars have successfully demonstrated the quantum advantages in some video processing tasks, but not concerning moving target segmentation. In this paper, a  quantum moving target segmentation algorithm
for grayscale video is proposed, which can use quantum mechanism to simultaneously calculate the difference of all pixels in all adjacent frames and then quickly segment out the moving target.  In addition,  a feasible quantum comparator
is designed to distinguish the grayscale values with the threshold. Then several quantum circuit units, including three-frame difference, binarization and AND operation, are designed in detail, and then are combined together to construct the complete quantum circuits for segmenting the moving target. For a quantum video with $2^m$ frames (every frame is a $2^n\times 2^n$ image with $q$ grayscale levels), the complexity of our algorithm can be reduced to O$(n^2 + q)$. Compared with the classic counterpart,  it is an exponential speedup, while  its  complexity is also superior to the existing quantum algorithms.
 Finally, the experiment is conducted on IBM Q to show the feasibility of our algorithm in the noisy intermediate-scale quantum (NISQ) era.

\end{abstract}


\section{Introduction}\label{sec1}

Compared with text, image and audio, video can contain more information and  convey information more intuitively, so the  video processing   is an important part in the development of computer vision in this video era. As one of the hot topics in video processing, moving target segmentation  plays an important role in computer vision, image applications, intrusion detection, abnormal behavior detection, etc. However, as the data of the video increases, the real-time problem gradually emerges.  Quantum computing can achieve amazing computing speed by virtue of its quantum advantage \cite{Nielsem,Abd-El-Atty}, so the combination of quantum computing and video processing can effectively solve the real-time problem.

How to store classical video into qubits is a primary problem in quantum video processing. Because each frame of the video is an image, we first need to represent each frame as a quantum image. The current quantum image representation methods mainly include the flexible representation of quantum image (FRQI) \cite{Le2010}, the multi-channel RGB images representation of quantum images (MCQI) \cite{Sun2013}, the normal arbitrary superposition state of quantum image (NASS) \cite{Li2014}, the quantum probability image encoding representation (QPIE) \cite{Yao2017}, and  the novel enhanced quantum representation (NEQR) \cite{Zhang2013}. 

Based on these quantum image representation models, some quantum video representation models have been proposed, such as the quantum movie (video) representation model based on FRQI \cite{Iliyasu2011}, the quantum video representation model based on MCQI model \cite{Yan2015} and the NEQR-based quantum video representation model QVNEQR \cite{Wang2016}. Among them, QVNEQR stores video information (frames and pixels) in the base state of qubits, which makes the operation of pixels in video simpler, so it has been widely used. At present, research on quantum video mainly focuses on quantum video representation \cite{Iliyasu2011, Yan2015,Wang2016}, moving target detection\cite{Wang2016,Yan2016M}, quantum video
stabilization \cite{Yan2016S}  and quantum video encryption \cite{Wang2017,Chen2018,Abd El-Latif2019,Song2020}.  

In 2015, Yan et al \cite{Yan2016M}. proposed a measurements-based moving target detection (MMTD) algorithm in a quantum video based on the measurement method, which can collapse the quantum video to a deterministic state, and then they observe the measurement results to judge the motion trajectory. This is also the only work to study quantum moving target detection algorithm to date. In the same year, they \cite{Yan2016S} proposed a quantum video stabilization strategy and they used quantum comparators and quantum image  translation operations to estimate the motion during exposure and to compensate for motion-induced video jitter. In 2020, Song et al. \cite{Song2020}  proposed a quantum video encryption scheme based on the new quantum video framework, the controlled-XOR operation and the improved logic mapping. The whole encryption process is performed by  inter-frame permutation, intra-frame pixel position geometric
transformation, and high 4-intra-frame-qubit-planes scrambling, which  has high efficiency with simple
calculation, low complexity and strongly filtrates.  All in all, the research of quantum video processing  is now in its initial stages, which may be attributed to the need to deal with the large amount of resources encoded in the frames. In the past decades, the quantum image segmentation, as an important part of computer vision, has made great progress \cite{Xia2019,Yuan2020,Zhang2015,Fan2019,Zhou2019,Chetia2021,Liu2022,Caraiman2014,Caraiman2015,Wang2022An,Wang2022A}, but the quantum video segmentation is still scarce.  In this paper, we use the  three-frame difference method to design a quantum counterpart and design a detailed quantum circuit. Unlike classical methods, quantum methods can take advantage of the parallelism of quantum computing to process the quantum superposition state pixels of the input video, which greatly speeds up the classical counterpart. 

In general, the main contributions of this work are as follows: 
\begin{itemize}
\item A moving target segmentation algorithm in quantum video is proposed, which can use quantum mechanism to simultaneously perform subtraction operation on all pixels in a grayscale video and quickly segment out the moving target.  
\item A quantum comparator with lower quantum cost is designed to distinguish the
grayscale values with the  threshold. Then, several specific quantum circuit units, including three-frame difference, binarization and AND operation, are designed in detail by using fewer qubits and quantum gates. And then based on these units, the complete quantum circuit is assembled to segment the moving target in quantum video. 
\item We verify the superiority and feasibility of our proposed algorithm by analyzing the circuit complexity and performing simulation experiment on IBM Q \cite{IBM}, respectively.
\end{itemize}

The rest of this paper is organized as follows. Section \ref{sec2}
introduces the principle of the NEQR model, QVNEQR model and the classical moving target segmentation algorithm. In Sect. \ref{sec3}, some basic quantum operation modules are introduced, then, a series of quantum circuit units are designed and some relevant quantum states equations are given. Section \ref{sec4} analyzes the circuit complexity of our algorithm and the experiment result. Finally, the conclusion is drawn in Sect. \ref{sec5}.

\section{Preliminaries}\label{sec2}

\subsection{Novel enhanced quantum representation (NEQR)}
In classical computers, information is stored and processed using binary numbers, and images are encoded in matrices that contain both position and color information \cite{Liu2018}. The NEQR \cite{Zhang2013} quantum image representation model entangles the grayscale value information and position information of each pixel in the classical image, and then uses a quantum superposition state to store a complete classical image. Assuming that the size of a classical image is $2^n\times 2^n$, and the grayscale range is $[0, 2^q-1]$, the grayscale value of each pixel needs to be represented by a binary sequence $c_{q - 1}^{}c_{q - 2}^{} \cdots c_0^{}$, $c_m\in \{0, 1\}(m = 0, 1,\cdots, q-1)$. $c_m $ represents each bit of a binary sequence, and $m $ represents the position of each bit. The specific form of the NEQR  model is as Eq.\ref{eq1}.

\begin{eqnarray}\label{eq1}
\left| I \right\rangle  = \frac{1}{{{2^n}}}\sum\limits_{i = 0}^{{2^{2n}} - 1} {\left| {{c_i}} \right\rangle \left| i \right\rangle }  = \frac{1}{{{2^n}}}\sum\limits_{i = 0}^{{2^{2n}} - 1} {\left| {c_{q - 1}^ic_{q - 2}^i \cdots c_0^i} \right\rangle \left| i \right\rangle },
\end{eqnarray}
where $i$ represents the position of the pixels and the position information of each pixel can be represented in the form shown in Eq. \ref{eq2}.
\begin{equation}\label{eq2}
\left| i \right\rangle  = \left| Y \right\rangle \left| X \right\rangle  = \left| {{Y_{n - 1}}{Y_{n - 2}} \cdots {Y_0}} \right\rangle \left| {{X_{n - 1}}{X_{n - 2}} \cdots {X_0}} \right\rangle, 
\end{equation}
where $Y$ and $X$ respectively represent the ordinate information and abscissa information of each pixel. $n-1,\cdots,1,0 $ represent the position of the n-bit binary sequence.

\subsection{Quantum video based on NEQR (QVNEQR)}

\begin{figure}
    \centering
    \includegraphics[width=6cm]{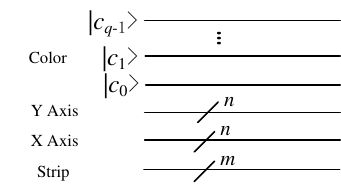 }
    \caption{Quantum circuit of QVNEQR.}
    \label{Fig1}
\end{figure}

The QVNEQR model \cite{Wang2016} is based on the NEQR model, and it combines multiple NEQR images to represent a quantum video, so each frame of the QVNEQR video is a NEQR image. Assuming that a video has $2^m$ frames, the size of each frame is $2^n\times2^n$, and the grayscale range is $[0, 2^q-1]$, then the QVNEQR model can be expressed as,
\begin{equation}
\left| V \right\rangle  = \frac{1}{{{2^{m/2}}}}\sum\limits_{j = 0}^{{2^m} - 1} {\left| {{I_j}} \right\rangle  \otimes } \left| j \right\rangle, 
\end{equation}

\begin{equation}
\left| {{I_j}} \right\rangle  = \frac{1}{{{2^n}}}\sum\limits_{i = 0}^{{2^{2n}} - 1} {\left| {{c_{j,i}}} \right\rangle  \otimes \left| i \right\rangle }  = \frac{1}{{{2^n}}}\sum\limits_{i = 0}^{{2^{2n}} - 1} {\left| {c_{q - 1}^{j,i}c_{q - 2}^{j,i} \cdots c_0^{j,i}} \right\rangle \left| i \right\rangle }, 
\end{equation}
where $c_s^{j,i} \in \left\{ {0,1} \right\}$, $s=0,1,\cdots, q-1$. The quantum circuit of a quantum video is shown in Fig. \ref{Fig1}, where $n$ and $m$ represent the  number of required qubits. If the video is a grayscale video, then $q=8$; if the video is a RGB video, then $q=24$.

\subsection{The classical moving target segmentation algorithm}

Moving target segmentation is to find out the moving objects in the video.  The most common methods are frame difference method \cite{Collins2000, Mahalingam2019, Tian2021}, background difference method \cite{Garcia-Garcia2020,Giraldo2022}  and optical flow method \cite{Kao2016}. Among them, the frame difference method is widely used because of its small calculation and high real-time performance. The video captured by the camera is a continuous sequence of images (frames), therefore, if there is no moving target in the video, the change from frame to frame is extremely weak; if there is a moving target in the video, the change from frame to frame will become very obvious. Based on this principle, the frame difference method performs a subtraction operation  on several consecutive frames in the video. That is, subtraction operations are performed between the pixels at the same position in different frames, and when the absolute value of the grayscale value difference exceeds a certain threshold, the pixel is determined as a part of the moving target. Commonly used frame difference methods include two-frame difference method and three-frame difference method, and the three-frame difference method combines the results of two adjacent frame difference methods, which can effectively avoid the ghosting problem of segmentation results and is widely used. Its mathematical expression is shown below.

\begin{equation}
  {d_1}(x,y) = \left\{ {\begin{array}{*{20}{c}}
{\begin{array}{*{20}{c}}
{1,}&{\left| {{f_k}(x,y) - {f_{k - 1}}(x,y)} \right|}
\end{array} \ge T}\\
{\begin{array}{*{20}{c}}
{0,}&{\left| {{f_k}(x,y) - {f_{k - 1}}(x,y)} \right| < T}
\end{array}}
\end{array}} \right.,  
\end{equation}

\begin{equation}
    {d_2}(x,y) = \left\{ {\begin{array}{*{20}{c}}
{\begin{array}{*{20}{c}}
{1,}&{\left| {{f_k}(x,y) - {f_{k + 1}}(x,y)} \right|}
\end{array} \ge T}\\
{\begin{array}{*{20}{c}}
{0,}&{\left| {{f_k}(x,y) - {f_{k + 1}}(x,y)} \right| < T}
\end{array}}
\end{array}} \right.,
\end{equation}

\begin{equation}
    D(x,y) = {d_1}(x,y) \cap {d_2}(x,y),
\end{equation}
where $(x, y)$ represents the position information of each pixel in the frames of the video (horizontal and vertical coordinates), $f_ {k-1} (x, y)$, $f_ k (x, y)$ and $f_ {k+1} (x, y)$ represent the $(k-1)$th, $k$th, and $(k+1)$th frames in the video $(k=2,3,4,5, \cdots)$. The corresponding $d(x, y)$ represents the image after the absolute value subtraction of adjacent frames, and $d(x, y)=1$ indicates that the pixel's value at $(x, y)$ in the obtained frame is 1, otherwise the pixel value at $(x, y)$ is 0. $T$ is the threshold value (it is artificially set according to the actual situation) and $D(x, y)$ represents the result frame in the video.

\section{The quantum moving target segmentation algorithm for grayscale video}\label{sec3}

\subsection{Quantum operations}
\begin{enumerate}[(1)]

\item  Quantum comparator
\\
The quantum comparator can compare the magnitude of two sequences of binary qubits, and it takes two qubit sequences  as input, and then outputs these two sequences $a$ and $b$ and their comparison result $y_1y_0$. Based on the idea of qantum bit string comparator (QBSC) \cite{Oliveira2007}, we designed a simpler quantum comparator (QC). In the process of comparison, we check bit by bit in order from high to low to see which binary number appears first as  1. If one of the bits in the same position of both numbers is  1 and the other is  0, the comparison is over and the result is output directly. Otherwise, the comparison will continue until the last bit. The detailed comparator quantum circuit is shown in Fig. \ref{Fig2}, where $a$ and $b$ denote the two numbers to be compared, $h$ denotes the auxiliary quantum bits, and $y_1y_0$ denotes the comparison result. If $y_1y_0$=10, then $a>b$; if $y_1y_0=01$, then $a<b$; if $y_1y_0=00$ or 11, then $a=b$. Compared with existing quantum comparators, our proposed comparator requires fewer quantum gates and auxiliary qubits, and the quantum cost is also lower, as shown in Table. \ref{tab1}.

\begin{table}[]
\centering
\caption{ Comparison of different quantum comparators.}\label{tab1}
\begin{tabular}{@{}cccc@{}}
\toprule
{ Quantum comparators} & { Auxiliary qubits} & { Quantum cost} &{ Distinguishable state}\\ \midrule
Reference\cite{Oliveira2007}                & 3$n$-1                       & 30$n$-15   &3    \\
Reference\cite{Li2020}                &1                       & 12$n$-8   &2    \\
Reference\cite{Yuan2020}                  & 5                          & 28$n$-15    &3   \\

Our comparator      & 3                          & 7$n$+6     &3     \\ \bottomrule
\end{tabular}
\end{table}

 \begin{figure*}
    \centering
    \includegraphics[width=10cm]{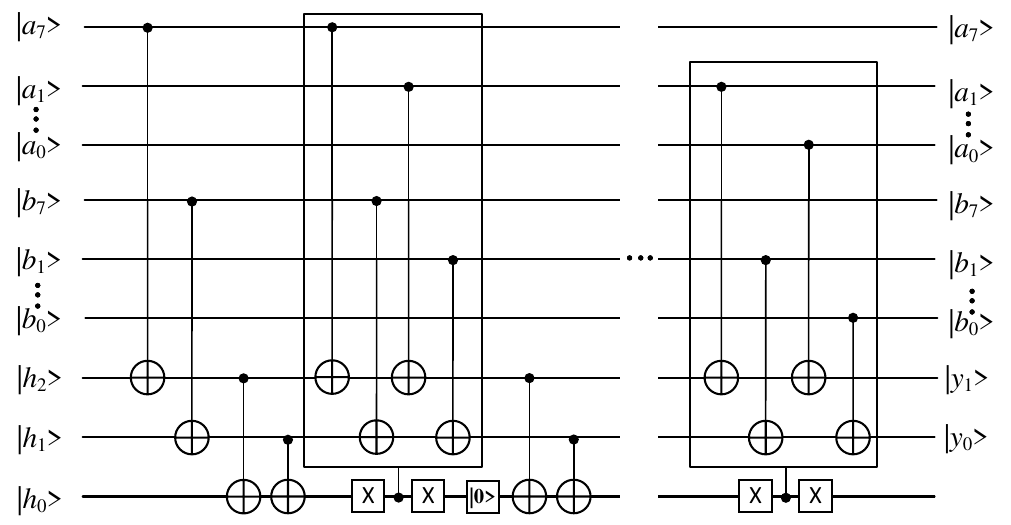}
    \caption{Quantum circuit realization of quantum comparator (QC).}
    \label{Fig2}
\end{figure*}

\item  Quantum subtractor
\\
 The quantum subtractor \cite{Wang2022A} we designed enables the subtraction operation of two binary qubit sequences $a$ and $b$, and it takes $a$ and $b$ as input, then takes the result of $a-b$  and $a$ as output. By  $a_0-b_0$, $a_1-b_1$, $\cdots $, $a_{n-1}-b_{n-1}$, the subtraction operation is performed bit by bit, and the borrowing information is reserved for the next bit, so that the subtraction operation can be completed using 3 auxiliary qubits, as shown in Fig. \ref{Fig3}. To prevent negative numbers during the calculation, we use a combination of a quantum comparator, a quantum subtractor, a Tofoli gate, four reset gates and $q$ CSWAP gates, as shown in Fig. \ref{Fig4}, so that it is always a larger number minus a smaller number.
 
 \begin{figure*}
    \centering
    \includegraphics[width=10cm]{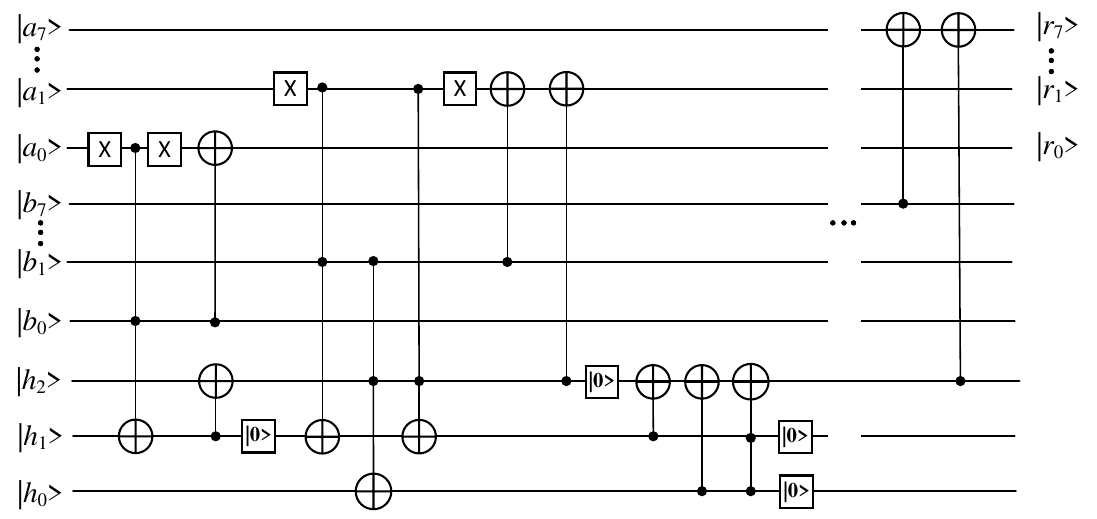 }
    \caption{Quantum circuit realization of quantum subtractor (QS).}
    \label{Fig3}
\end{figure*}

 \begin{figure}
    \centering
    \includegraphics[width=8cm]{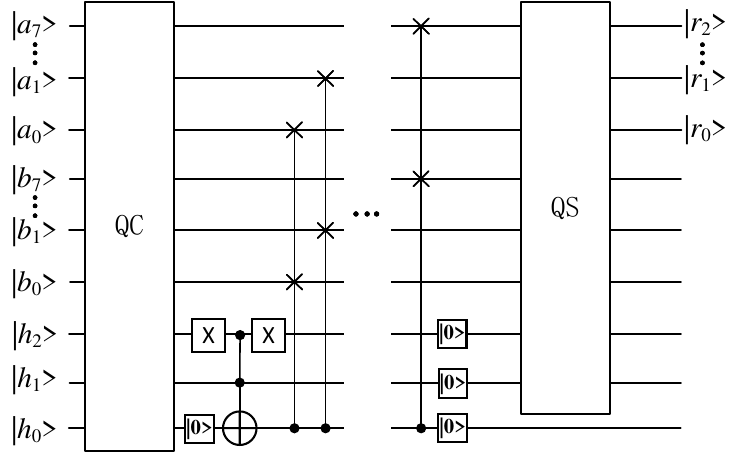  }
    \caption{Quantum circuit realization of quantum absolute value subtractor (QAS).}
    \label{Fig4}
\end{figure}

\item Quantum copy operation
\\
In order to make a copy of the original video, we need to use the copy operation \cite{Iliyasu2013} to do so. It is implemented by several CNOT gates and auxiliary qubits. The specific quantum circuit is shown in Fig. \ref{Fig5}, where $  | x \rangle$ represents the original value and $| 0\rangle$ is used to store the copied value.

\begin{figure}
    \centering
    \includegraphics[width=9cm]{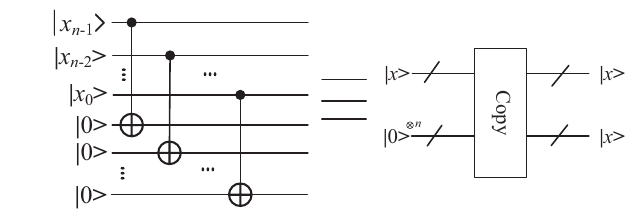 }
    \caption{Quantum circuit realization of copy.}
    \label{Fig5}
\end{figure}

\item Cycle shift transformation operation
\\
The cycle shift transformation (CT) operation \cite{Iliyasu2011} between frames in quantum video allows all frames to be shifted cyclically so that every frame in the video can get its neighboring frames at the same time. For example, if we shift the video forward by one unit (CTs-), then the k-th frame in the video becomes the (k-1)-th frame. The unitary operation for cyclic shifting of a quantum video containing $2^m$ frames is shown below and the specific quantum circuit is shown in Fig. \ref{Fig6}.

\begin{equation}
    CT(j \pm )\left| V \right\rangle  = \frac{1}{{{2^{m/2}}}}\sum\limits_{j = 0}^{{2^m} - 1} {\left| {{I_j}} \right\rangle  \otimes } \left| {(j \pm 1)\bmod {2^m}} \right\rangle. 
\end{equation}

\begin{figure}
    \centering
 \subfigure[]{\includegraphics[width=6cm]{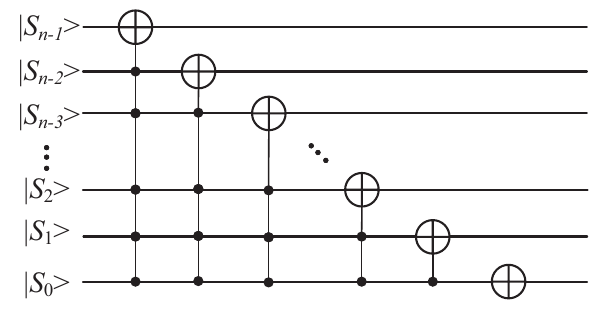}}
 \subfigure[]{\includegraphics[width=6cm]{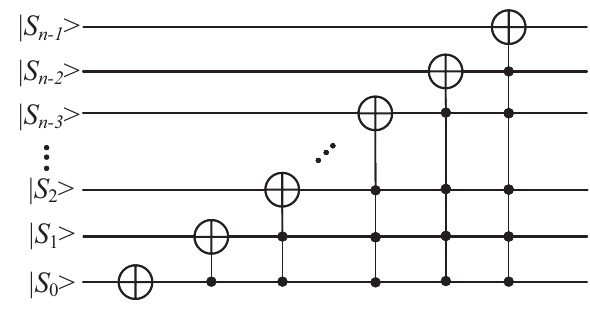}}
    \caption{The schematic diagram of the X-axis cyclic shift transformation.}
    \label{Fig6}
\end{figure}

\end{enumerate}
\subsection{The proposed algorithm}
\begin{figure*}
    \centering
    \includegraphics[width=14cm]{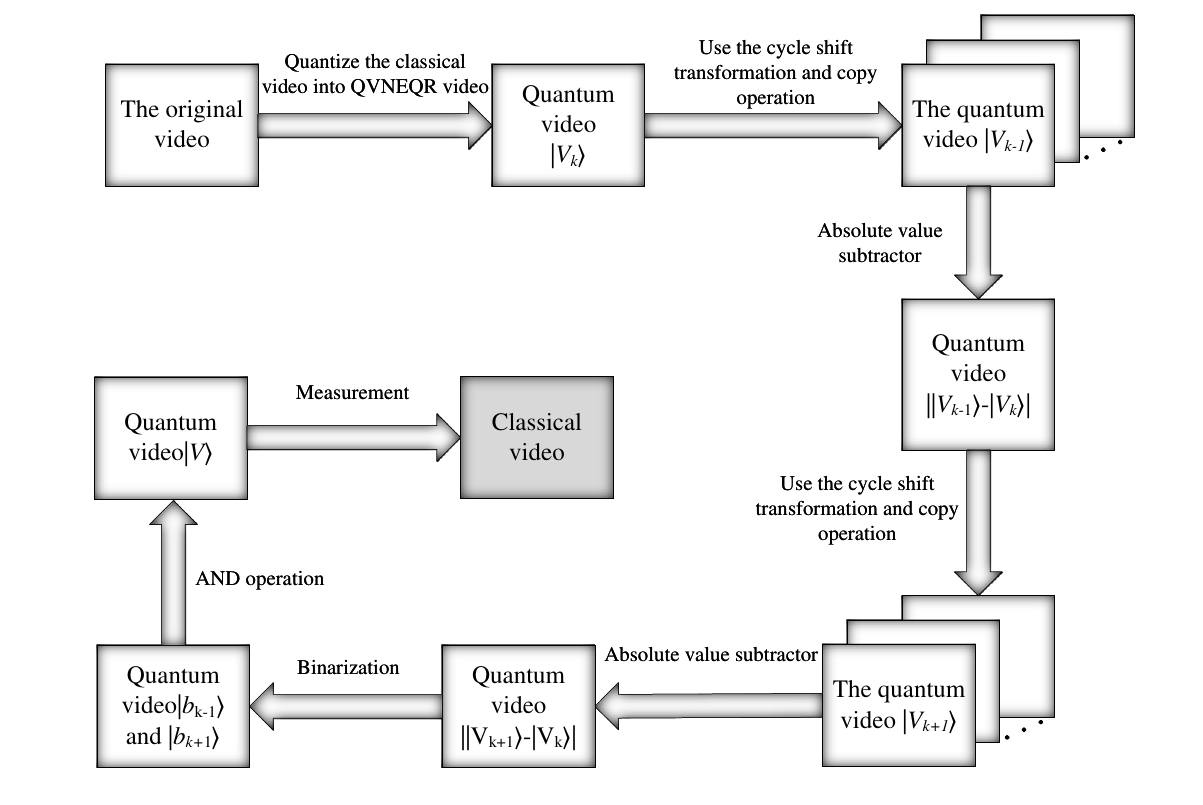 }
    \caption{The workflow of our algorithm.}
    \label{Fig7}
\end{figure*}

 In order to segment the moving targets in quantum video, our proposed segmentation algorithm can be described as  follows (also shown in Fig. \ref{Fig7}).

 \begin{enumerate}[Step 1:]
     \item  Quantum Video $ \left| V_k\right\rangle$ is prepared based on the QVNEQR model,
           ${V_K}\xrightarrow{{QVNEQR}}\left| {{V_K}} \right\rangle$.
     
     \item The copy operation is used to copy $V_k$ and store it in the auxiliary qubits for backup, \\ $\left| {{V_K}} \right\rangle {\left| 0 \right\rangle ^{ \otimes q}}\xrightarrow{{copy}}\left| {{V_K}} \right\rangle \left| {{V_K}} \right\rangle $.
     
     \item The original video is cycle shifted forward by one frame (from frame $k$ to frame $ k-1$) by using a cycle shift transformation to obtain a new video $\left | V_{k-1} \right\rangle$,  $\left| {{V_K}} \right\rangle \xrightarrow{{CT}}\left| {{V_{K - 1}}} \right\rangle$.
     
     \item  The quantum absolute value subtraction is used on the above two videos ($\left| V_{K - 1} \right\rangle  $ and $\left| V_k \right\rangle $),  i.e. absolute value subtraction for each pixel of each frame, $\left| {\left| {{V_{K - 1}}} \right\rangle  - \left| {{V_K}} \right\rangle } \right| = \left| {{D_{K - 1}}} \right\rangle$.
     
     \item The copy operation is used to reset $ \left| V_k\right\rangle$, $\left| {{V_K}} \right\rangle {\left| 0 \right\rangle ^{ \otimes q}}\xrightarrow{{copy}}\left| {{V_K}} \right\rangle \left| {{V_K}} \right\rangle $.
     
     \item A new video $\left| V_{k+1} \right\rangle$ is obtained by shifting the original video backwards by one frame (from frame $k$ to frame $k+1$) using a cycle shift transformation,  $\left| {{V_K}} \right\rangle \xrightarrow{{CT}}\left| {{V_{K + 1}}} \right\rangle$.

     \item The quantum absolute value subtraction is used on the above two videos ($\left| V_{k+1}\right\rangle$ and $\left|V_k\right\rangle$), i.e. absolute value subtraction for each pixel of each frame, $\left| {\left| {{V_{K + 1}}} \right\rangle  - \left| {{V_K}} \right\rangle } \right|=\left| {{D_{K + 1}}} \right\rangle$.
     
     \item The result videos after two subtractions are binarized simultaneously, $\left| {{D_{K \pm 1}}} \right\rangle \xrightarrow{{QB}}\left| {{b_{K \pm 1}}} \right\rangle$.

     \item The two videos after binarization are subjected to the AND operation and the final result quantum video is obtained, $\left| {{b_{K - 1}}} \right\rangle AND\left| {{b_{K + 1}}} \right\rangle \xrightarrow{{}}\left| V \right\rangle$.
     
     \item The quantum measurement operation is performed, then the classical image is retrieved and the algorithm ends, $\left| V \right\rangle \xrightarrow{{Measurement}}V$.
     
 \end{enumerate}

\subsection{The quantum circuit implementation of the algorithm}
 \begin{enumerate}[(1)]
 
 \item Quantum video preparation
 \\
 If a video has $2^m$ frames, and each frame is an image of size $2^n\times2^n$ with grayscale level $q$, then according to the QVNEQR model, we need $2n$ qubits to store the position information and $q$ qubits to store the grayscale value information of the pixels in each frame, besides,  $m$ qubits are also needed to store the frame information. Since we need to perform the frame shift operation on the quantum video, the obtained video and the original video share the position information, so we also need  additional $2q$ qubits to store the obtained video ($V_{k-1}$ and $V_{k+1}$) and to copy the original video ($V_k$). The quantum state expression is shown below.
 
  \begin{equation}
     \left| V \right\rangle  \otimes {\left| 0 \right\rangle ^{2q}} = \frac{1}{{{2^{\frac{m}{2} + n}}}}\sum\limits_{j = 0}^{{2^m} - 1} {\sum\limits_{i = 0}^{{2^{2n}} - 1} {\left| {{c_{j,i}}} \right\rangle \left| i \right\rangle } } \left| j \right\rangle {\left| 0 \right\rangle ^q}{\left| 0 \right\rangle ^q}.
 \end{equation}
 
 \item Video frame cycle shift
 \\
 Since all pixels in all frames of the video are in the superposition state, the processing of a pixel is also the processing of all pixels in the video. So we shift the video one frame forward and one frame backward respectively, so that we have three videos ($V_{k-1}$, $V_k$, $V_{k+1}$) , and the corresponding frames $K$ are the three adjacent frames of the original video (frame $k-1$, $k$, $k+1$). This allows us to operate on all frames at once when we perform the frame difference operation on a particular frame. The expression for the quantum state of the quantum video set after the cycle shift is shown as follows.

 \begin{equation}
     \frac{1}{{{2^{\frac{m}{2} + n}}}}\sum\limits_{j = 0}^{{2^m} - 1} {\sum\limits_{i = 0}^{{2^{2n}} - 1} {\left| {{c_{j - 1,i}}} \right\rangle \left| {{c_{j,i}}} \right\rangle \left| {{c_{j + 1,i}}} \right\rangle \left| i \right\rangle } } \left| j \right\rangle. 
 \end{equation}
 
 \item Frame difference operation
 \\
After shifting the video, we have to perform an absolute value subtraction operation (difference operation) on these three corresponding frames of the video. This is also known as $|V_{k-1}-V_{k}|$ and $|V_{k+1}-V_k|$. The operation specific to each pixel is the absolute value subtraction of the pixels at the same position in the k-th frame of the three videos. Since all pixels in the video are in the superposition state, our operation on one pixel is also an operation on all pixels. The specific quantum circuit is shown in Fig. \ref{Fig8}.

\begin{figure}
    \centering
    \includegraphics[width=9cm]{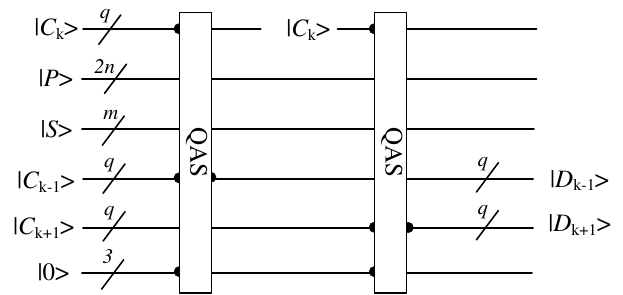  }
    \caption{Quantum circuit realization of Frame Difference.}
    \label{Fig8}
\end{figure}
 
 \item Binarization
 \\
 In order to be able to reduce the amount of computation and quickly distinguish the pixel points that have changed, we need to binarise the two images after the frame difference operation. First, we compare the pixels to the threshold $T$ by using a quantum comparator.  If $c \geq T$, then the result of the comparator is $y_1y_0\neq01$. At this point, if $c_0=0$, then $c_0$ is set to 1 by using the CNOT and Toffoli gates. If the result of the comparator is $y_1y_0=01$, this means that $c<T$. At this point, if $c_0=1$, then $c_0$ is set to 0 by using the CNOT and Toffoli gates. The specific quantum circuit is shown in Fig. \ref{Fig9}.
 
 \begin{figure}
    \centering
    \includegraphics[width=8cm]{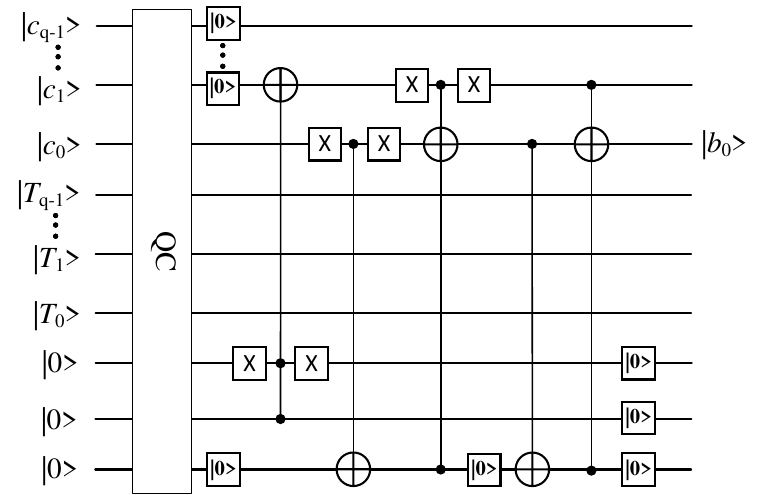  }
    \caption{Quantum circuit realization of quantum binarization (QB).}
    \label{Fig9}
\end{figure}
 
 \item AND operation
 \\
 After the above operations, we get two binary videos which contain the moving targets between two adjacent frames, however, when the target is moving more slowly, the ghosting will occur. Therefore, we perform the AND operation on the two binary videos to accurately segment the moving targets, and we use the Toffoli gate for this operation. This means that we decide that two targets segmented from the video are moving targets only when they overlap. The specific quantum circuit is shown in Figure. \ref{Fig10}, and the complete quantum algorithm circuit is shown in Figure. \ref{Fig11}.
 
 \begin{figure}
    \centering
    \includegraphics[width=3cm]{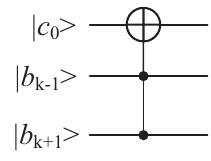  }
    \caption{Quantum circuit realization of  AND operation.}
    \label{Fig10}
\end{figure}

  \end{enumerate}
  
  \begin{figure*}
    \centering
    \includegraphics[width=13cm]{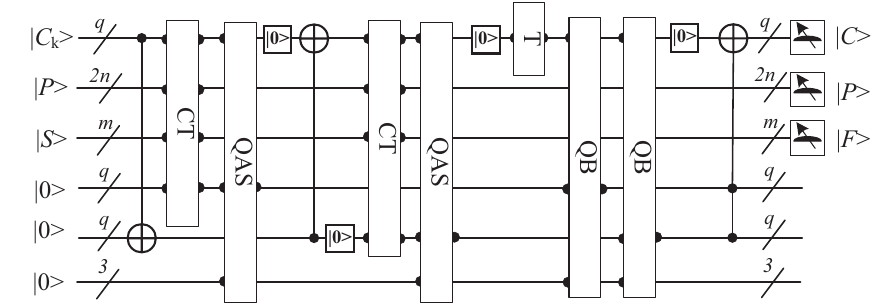  }
    \caption{Quantum circuit realization of  the Complete quantum algorithm. }
    \label{Fig11}
\end{figure*}

\begin{figure*}
\centering
    \subfigure[]{
   \includegraphics[width=3cm]{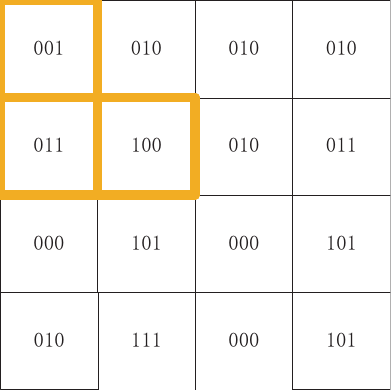}}
   \subfigure[]{
  \includegraphics[width=3cm]{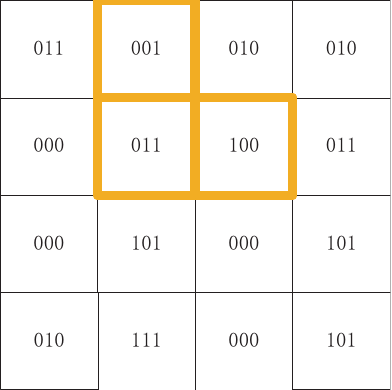}}
   \subfigure[]{
  \includegraphics[width=3cm]{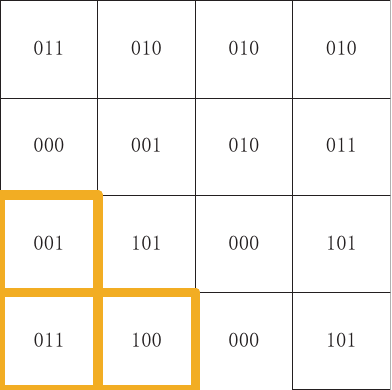}}
   \subfigure[]{
  \includegraphics[width=3cm]{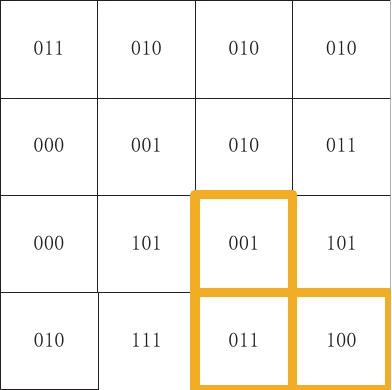}}
\caption{The video including a moving target.(a,b,c and d  are four consecutive frames in the video.}
\label{Fig12}
\end{figure*}

\begin{figure*}
    \centering
    \includegraphics[width=20cm, height=10cm, angle=270]{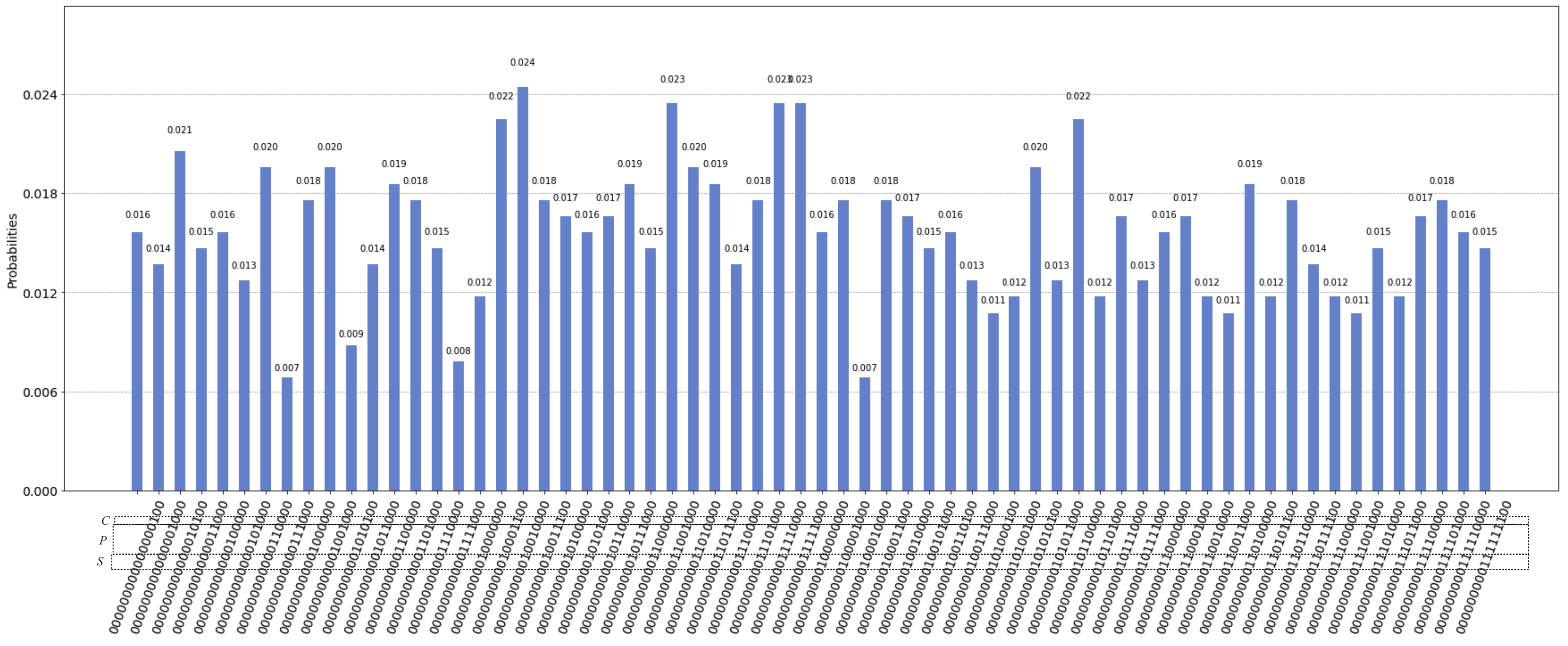 }
    \caption{The probability histogram of the resulting video}
    \label{Fig13}
\end{figure*}

\begin{figure*}
\centering
 \subfigure[]{
\includegraphics[width=3cm]{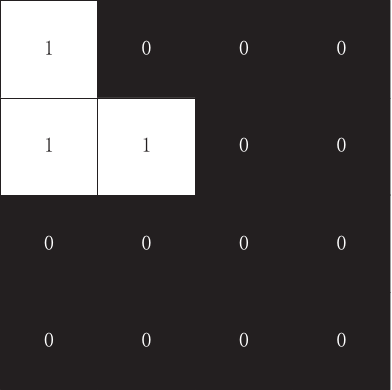}}
 \subfigure[]{
\includegraphics[width=3cm]{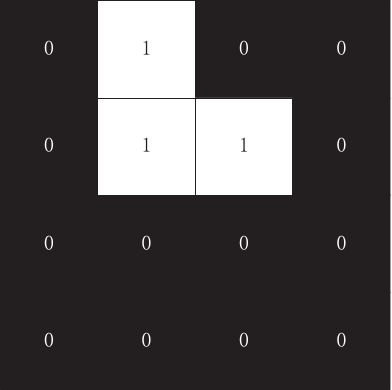}}
 \subfigure[]{
\includegraphics[width=3cm]{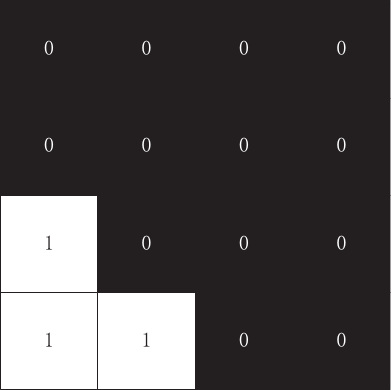}}
 \subfigure[]{
\includegraphics[width=3cm]{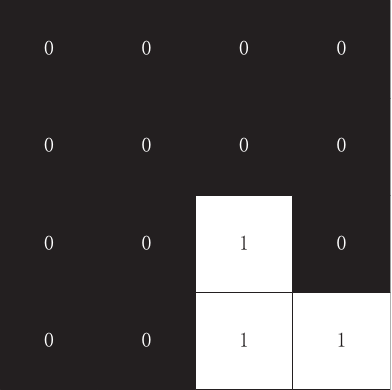}}
\caption{ The result video processed by our proposed quantum algorithm. (a,b,c and d  are four consecutive frames in the video, which include a moving target maked with 1.)}
\label{Fig14}
\end{figure*}

\begin{figure*}[!t]
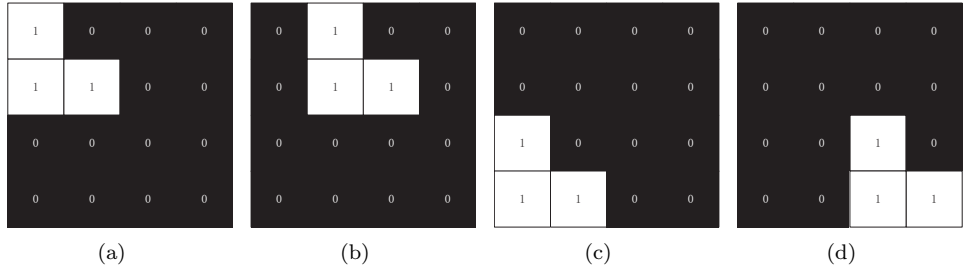

\centering
 \subfigure[]{
\includegraphics[width=3cm]{Fig15.1}}
 \subfigure[]{
\includegraphics[width=3cm]{Fig15.2}}
 \subfigure[]{
\includegraphics[width=3cm]{Fig15.3}}
 \subfigure[]{
\includegraphics[width=3cm]{Fig15.4}}
\caption{The result video processed by the classical algorithm. (a,b,c and d  are four consecutive frames in the video, which include a moving target maked with 1.)}
\label{Fig15}
\end{figure*}

\section{Circuit complexity and experiment analysis}\label{sec4}

\subsection{Circuit complexity analysis}

The proposed algorithm uses a total of $q+2n+m+q+q+3$ qubits, as shown in Figure \ref{Fig11}. $C_k$ represents the grayscale value of each pixel in the original video. Because the pixel qubits in the video are entangled with position qubits, only $q$ qubits are required for a video with a grayscale value of $[0, 2^{q-1}]$. $P$ represents the position information of each frame’s pixels in the original video, and the qubits are in superposition state. Therefore, for all frames with size $2^n \times 2^n$,  only $2n$ qubits are required to represent the position information of all pixels in all frames. $S$ represents the position information of the frame, and its qubits are also in superposition state. Therefore, for a video containing $2^m$ frames, only $m$ qubits are required to represent it. The remaining qubits are used as auxiliary qubits to replicate the original video and form the computational unit. Each replication only requires $q$ qubits to store the grayscale value information, as pixel position information and frame position information share qubits with the original video.

The complexity of a quantum circuit is determined by the number of underlying logic gates in the circuit, and the entire quantum circuit for processing quantum video is made up of several quantum logic gates. Each quantum gate can be composed of single or double qubit gates. Therefore, we consider the complexity of a single and double qubit gate as unit 1, so a NOT gate, a CNOT gate, and a reset gate all have a complexity of 1. A Toffoli gate can be composed of 5 two-qubit gates, so its quantum cost is 5 \cite{ Nielsem}, and a CSWAP gate has a complexity of 3 \cite{Li2020}. Supposing a quantum video has $2^m$ frames, each of which has a size of $2^n \times 2^n$ and a grayscale range of $[0,2^{q-1}]$. Therefore, we discuss the complexity of the quantum circuit in the following parts.

Since there are no devices to obtain quantum video directly nowadays, we need to convert classical video into quantum video, which consumes a lot of quantum resources. But in quantum video processing or quantum image processing, we usually ignore the complexity of this part  because our algorithm is for quantum video. This is just like classical algorithms which also do not take into account the complexity of video preparation when calculating the algorithm complexity. Therefore, the complexity of this stage is 0.

 After the quantum video has been prepared, we need to copy the video into the auxiliary qubits by using the copy operation and recover the video after the adjacent frame difference. The copy operation is composed of $q$ CNOT gates, so its complexity is O($q$).

 In the algorithm we need to perform two CT operations on the video, which include the forward and backward shift of each frame in the video. The CT operation consists of a series of Toffoli gates with multiple control qubits, so this part has a complexity of O$(n^2)$ \cite{Li2020,Li2019}.

The quantum absolute value subtraction (QAS) between two adjacent frames is required in our algorithm and each frame difference is performed by an absolute value subtractor.  Each QAS is composed of a quantum comparator, a quantum subtractor, a Toffoli gate, four reset gates and $q$ CSWAP gates and their complexity is O($q$),O($q$), 5, 4 and O($q$) respectively, so that the complexity of the QAS operation is O($q+q+5+4+q$)=O($q$).

 Since the threshold is a known value, we set it using $q$ reset gates and $q$ NOT gates. The complexity of this part is O($q+q$)=O($q$).
 
 The binarization operation  requires a quantum comparator, two CNOT gates, three Toffoli gates and $q+4$ reset gates, which have a complexity of O($q$), 2, 15, O($q+4$) respectively; hence the complexity of this part is O($q+2+15+q+4$)= O($q$).
 
 The AND operation is implemented by using a Toffoli gate with a complexity of 5 (O(1)). In order to compose the complete quantum circuit, we need to use $4q$ reset gates  at different places, which has the complexity of O($q$).

In summary, the complexity of our proposed quantum video segmentation algorithm is O($q+n^2+q+q+q+1+q$)=O($n^2+q$). The classical video segmentation algorithms need to process each pixel of each frame individually, so the complexity is no less than O($2^{2n+m}$). Therefore, in comparison, the performance of our algorithm can reach exponential improvement, which can be a good solution to the real-time problem in the classical algorithm, as can be seen in  Table \ref{Tab2}. 

 In addition, the existing moving  target detection algorithm MMTD in quantum video \cite{Yan2016M} is based on measurement. It uses MCQI model to encode quantum image, and the complexity of image preparation is O$(2^{4n})$, where $n$ is the number of quantum bits required. In order to detect moving targets in quantum video with $2 ^ m$ frames, the complexity of the algorithm is O$(q\cdot2^{m+4n})$. If the video preparation process is not considered, the complexity is O$(q\cdot2^m)$, which is much higher than the complexity of our algorithm. Since the development of segmentation algorithms in quantum video is still immature, there are fewer related works. And the video is composed of many images, therefore, we choose some quantum image segmentation algorithms for comparison. For convenience, we abbreviate the segmentation algorithms in  Ref. \cite{Fan2019} and Ref. \cite{Chetia2021} as EDCS and EDFS. EDCS employs the classical Sobel operator for edge detection thereby segmenting the target in the image. It requires gradient computation and threshold comparison for each pixel, thus the complexity is O($n^2+2^{q+4}$). While  EDFS uses an improved Sobel operator to detect pixel gradients in more directions. In addition, non-maximum suppression technique and double threshold detection technique are also added into the algorithm so that the edges of the target can be segmented more accurately. So the complexity of the algorithm is O($n^2+q^3$). From the above comparison, it can be seen that even though our algorithm deals with video with $m$ frame images, its complexity is much lower than the image segmentation algorithms, which is very meaningful in this NISQ era.

\begin{table}[]
\center
\caption{Complexity comparison of different algorithms.}\label{Tab2}
\begin{tabular}{@{}lll@{}}
\toprule
Algorithm       &Application Scenario   & Complexity \\ \midrule
Classic algorithm\cite{Tian2021}  &Video  & O($2^{2n+m}$)           \\
MMTD\cite{Yan2016M}     &Video          & O($q\cdot 2^{m}$)           \\
EDCS\cite{Fan2019} &Image  &O($n^2+2^{q+4}$)\\
EDFS\cite{Chetia2021} &Image  &O($n^2+q^3$)\\
Our algorithm  &Video  & O($n^2+q$)           \\ \bottomrule
\end{tabular}
\end{table}

\subsection{Experiment}

To verify the practical feasibility  of the quantum algorithm, we used the quantum cloud platform provided by IBM Q (IBM Quantum Experience) \cite{IBM}. The environment created with the Qiskit \cite{Aleksandrowicz2019} extension package and anaconda compiles the Python language into the OpenQASM language, which can then create quantum circuits, but because of the preciousness of qubit, we can only use a quantum computer with 5 qubits, which is far from enough. So we can only use the quantum simulator  “ibm-qasm-simulator”     to run our quantum circuits, which has 32 qubits and supports all the quantum logic gates we use. For every task, the run time of this simulator is limited to 10,000 second, and when  used, it is mainly called by the statement ``backend=IBMQ.get\_provider(`ibm-q').get\_backend(`ibmq\_qasm\_simulator')".

 For the currently available experimental conditions, we have chosen a video containing $2^2$ frames with a fixed background and each frame has a size of $2^2\times 2^2$ and a grayscale range of $[0,2^3-1]$, as shown in Fig. \ref{Fig12}. The pixels marked in the figure are the moving targets, and the grayscale of each pixel is marked in the figure in binary form.

We use the proposed algorithm to segment the moving targets in the video, and because of the small range of grayscale values in the video, we set the threshold value to 001, which can be adjusted according to the segmentation effect in practice. In addition, to save experimental time, we only measured the key qubits (the number of measurements is 1024), i.e. the frame position $S$  of the video, the pixel position $P$ of each frame and the grayscale value $C$. The measurement results are shown in Fig. \ref{Fig13}. The horizontal coordinates of the graph indicate the specific information of each pixel in the video and some auxiliary qubits. The vertical coordinates indicate the measured probability of each  qubit sequence. Based on the probability histogram information, we plotted the segmented quantum video as shown in Fig. \ref{Fig14}. The pixels with a grayscale value of 1 are the segmented moving targets. As can be seen from the figure, our algorithm can accurately segment the moving targets in the video.

Figure \ref{Fig15} shows the video  processed by the classical algorithm, from which it can be seen that the processing effect of the classical algorithm is the same as that of our quantum algorithm. This is because our quantum algorithm utilizes quantum mechanisms to accelerate classical algorithms and does not change the way classical algorithm processes the video. Therefore, our algorithm's processing results is the same as that of classical algorithm. However, our algorithm has an exponential improvement in complexity, which can solve the real-time problem that occurs with the classical algorithm without affecting the segmentation results.

\section{Conclusion}\label{sec5}

Although the  classical  moving target segmentation algorithms in video are mature, how to improve their processing efficiency by using quantum mechanisms is still in its infancy in this video era.
  In this paper, a quantum moving target segmentation algorithm for grayscale video is proposed, which can use quantum mechanisms to process all pixels in the video  at the same time and segment out the moving target. In addition,  a quantum comparator with lower quantum cost is designed  and a complete circuit for quantum video segmentation is given. The circuit complexity  and experiment analysis demonstrate the superiority and feasibility of our algorithm.

When segmenting the moving target in a  video, our algorithm is only adapted to the video with a constant background, such as surveillance. However, in practical applications, there are many scenarios that are variable, so it is our future research direction to segment the video containing changing background.

\medskip
\textbf{Acknowledgements} \par 
This work is supported by the National Natural Science Foundation of China (62071240), the Innovation Program for Quantum Science and Technology (2021ZD0302900), and the Priority Academic Program Development of Jiangsu Higher Education Institutions (PAPD).

\medskip
\textbf{Conflict of Interest} \par
The authors declare no conflict of interest.

\medskip
\textbf{Data Availability Statement} \par
All data generated or analysed during this study are included in this published article
[and its supplementary information files].

\medskip

%


\end{document}